\documentclass[letterpaper, 10 pt, conference]{ieeeconf}
\IEEEoverridecommandlockouts
\usepackage{cite}
\usepackage{amsmath,amssymb,amsfonts}
\usepackage{booktabs}
\usepackage{balance}
\usepackage{algorithmic}
\usepackage{tabularx}
\usepackage{caption}
\usepackage{graphicx}
\usepackage{textcomp}
\usepackage{xcolor}
\usepackage{dsfont}

\begin{document}

\title{
A Multimodal Neural Network for Recognizing Subjective Self-Disclosure Towards Social Robots}

\author{Henry Powell$^{1,2*}$, Guy Laban$^{3,4,2*}$, Emily S. Cross$^{2,5}$\\
\makebox[\linewidth][c]{%
  \parbox{\linewidth}{\centering
  \normalsize  $^{1}$Amazon, Edinburgh, UK.\\
    $^{2}$School of Psychology and Neuroscience, University of Glasgow, Glasgow, UK.\\
    $^{3}$Department of Industrial Engineering and Management, Ben-Gurion University of the Negev, Beer Sheva, Israel.\\
    $^{4}$Department of Computer Science and Technology, University of Cambridge, Cambridge, UK.\\
    $^{5}$Social Brain Sciences Group, Department of Humanities, Social and Political Sciences, ETH Zurich, Zurich, Switzerland.
  }%
}%
\thanks{*Henry Powell and Guy Laban have contributed equally to this work and share first authorship. Corresponding author: guy.laban@cl.cam.ac.uk
}
\thanks{The authors gratefully acknowledge funding from the European Research Council (ERC) under the European Union’s Horizon 2020 research and innovation programme (Grant agreement 677270 to E.S.C.), the Leverhulme Trust (PLP-2018-152 to E.S.C.), and the European Union’s Horizon 2020 research and innovation programme under the Marie Sklodowska-Curie to ENTWINE, the European Training Network on Informal Care (grant agreement no. 814072 to E.S.C.). Our thanks also go to Hubert Ramsauer for helpful discussions on the use of Hopfield networks.}}

\maketitle

\begin{abstract}
Subjective self-disclosure is an important feature of human social interaction. While much has been done in the social and behavioural literature to characterise the features and consequences of subjective self-disclosure, little work has been done thus far to develop computational systems that are able to accurately model it. Even less work has been done that attempts to model specifically how human interactants self-disclose with robotic partners. It is becoming more pressing as we require social robots to work in conjunction with and establish relationships with humans in various social settings. In this paper, our aim is to develop a custom multimodal attention network based on models from the emotion recognition literature, training this model on a large self-collected self-disclosure video corpus, and constructing a new loss function, the scale preserving cross entropy loss, that improves upon both classification and regression versions of this problem. Our results show that the best performing model, trained with our novel loss function, achieves an F1 score of 0.83, an improvement of 0.48 from the best baseline model. This result makes significant headway in the aim of allowing social robots to pick up on an interaction partner's self-disclosures, an ability that will be essential in social robots with social cognition.

\end{abstract}

\section{Introduction}
Self-disclosure is the sharing of one's thoughts, feelings, or personal information during a social interaction \cite{RefWorks:509}. It is an import facet of human social behaviour and can contribute to many aspects of our lives. It can contribute to the extent to which we form bonds with one another - i.e. how intimate and important we consider our relationship with others to be - as well as contributing significantly to our mental and physical health \cite{RefWorks:513, RefWorks:509, RefWorks:357}. In this paper, we focus specifically on \textit{subjective self-disclosure}, which picks out the degree to which one \textit{perceives} themselves to be sharing personal information with others. For example, it may be the case that someone shares some information which may, in general, not be perceived to be particularly personal or sensitive. However, this information might well be meaningful to the person disclosing it. Here the term subjective is supposed to clarify that what is important in an act of self-disclosure is that the person performing it believes themselves to be sharing meaningful thoughts, feelings, or personal information.

Considering the importance of subjective self-disclosure in developing meaningful personal relationships, we argue that sensitivity to these self-disclosures becomes a crucial attribute of social robots when establishing relationships with them. Social robots are increasingly being studied in social contexts \cite{Henschel2021}, with a focus on their affective roles \cite{Churamani_adapt_2020,
Spitale2025PastWell-being}.Previous research has highlighted how individuals develop social relationships with these agents over time \cite{Laban2024BuildingTime,Laban2025CopingCaregiversb}, why users open up to them \cite{Laban2023OpeningBehavior,Laban2024SharingFeel}, and the diverse ways in which they can support human users \cite{Laban2024BuildingTime,RefWorks:404,Laban2024,Laban2025CopingCaregiversb,laban_lon}. Therefore, these robotic agents, designed to operate among humans and communicate with them, must be equipped to handle such disclosures appropriately \cite{Laban2024SharingFeel}. However, despite their potential, social robots currently lack the nuanced cognitive capacity to infer such social information and understand subjective perceptions as effortlessly as humans do \cite{RefWorks:426}, which is critical when these robots take social and affective roles. Moreover, there is very little, if any, work in the field of human--robot interaction (HRI) that seeks to model the ability to detect and measure self-disclosure with the aim bestowing a socially oriented artificial agent with this ability. One such exception is a previous study from our group, where we found that a number of standard deep learning architectures were able to perform well above average on the task of ranking the degree of users' subjective self-disclosure in recorded interactions \cite{Powell2022}.


In this study, we aim to address this problem and to improve previous results. We did this in a number of ways. Firstly, by developing a significantly larger data set that included an visual as well as an audio modality in order to capture markers of subjective self-disclosure that may be present in how facial features evolve over time. Secondly, by developing a more sophisticated deep learning model that was better suited to the task i.e. one that was developed using domain knowledge of the problem and the data representations we used as input to our model. Thirdly, to address the problem of experimental framing that we experienced in that study, i.e. how to model data that was both categorical and scaled.

\subsection{The Current Paper}
The remainder of the paper takes the following form: in Section \ref{DataSetCollection} we detail the design, data collection and data pre-processing for the experiment that we conducted in order to form the dataset used to perform our deep learning experiments. Next, in Section \ref{FeaturesExtraction} we outline which features we extracted from the processed dataset and the means by which we extracted them. In Section \ref{DeepLearningExperiments}, we describe the architecture of our multi modal attention network in detail. We then describe the experiments we conducted to produce baselines to which we could compare the performance of this model. Further, we detail the parameters of our ablation experiment to test the effects of the loss functions, feature sets, and experimental framings that we used, and finally, the specific details of the training implementation. Then, in Section \ref{Results} we present the results of the ablation study before finally, in Section \ref{DiscussionConclusion}, discussing some areas for further improvement to our approach and some issues with it.

\subsection{Our Contribution}

Our contributions to the field 
are as follows:
\begin{enumerate}
    \item We present the most extensive attempt to model subjective self-disclosure in HRI so far,
    \item A multi modal attention based architecture designed specifically for self-disclosure modelling from audio and video data,
    \item A novel loss function, the scale preserving cross entropy loss, that effectively deals with problems that fall between regression and classification and outperforms both squared error and cross entropy approaches to self-disclosure modelling.
\end{enumerate}

\section{Data Set and Data Collection}\label{DataSetCollection}

In order to generate data for the models, a long-term mediated online experiment was conducted, as reported in \cite{Laban2024BuildingTime}. We repeat that protocol here verbatim for consistency: A 2 (Discussion Theme: COVID-19 related vs. general) by 10 (chat sessions across time) between-groups repeated measures experimental design was followed. 39 Participants were randomly assigned to one of the two discussion topic groups, according to which they conversed with the robot Pepper (SoftBank Robotics) via Zoom video chats about general everyday topics (e.g., social relationships, work-life balance, health and well-being). One group’s conversation topics were framed within the context of the COVID-19 pandemic (e.g., social relationships during the pandemic), whereas the other group’s conversation topics were similar, except that no explicit mention of the COVID-19 pandemic was ever made. Participants were scheduled to interact with the robot twice a week during prearranged times for five weeks, resulting in 10 interactions in total. Each interaction consisted of the robot asking the participant 3 questions (x3 repetitions), starting with a generic question to build rapport (e.g., how was your week/weekend), followed by two additional questions that corresponded to one of the 10 randomly ordered topics (for the topics, questions, and examples see \cite{Laban2024BuildingTime}). 
The topic of each interaction was assigned randomly, as was the order of the questions. After conversing with Pepper via the zoom chat, participants filled a questionnaire reporting for their perceptions of their subjective disclosure via an adapted version of Jourad self-disclosure questionnaire \cite{RefWorks:509}. The zoom chats were recorded for analysis purposes. Each interaction with the robot lasted between 5 to 10 minutes, and another 10-20 minutes were taken up completing questionnaires. The study followed rigorous ethical standards and all study procedures were approved by the research ethics committee of the School of Psychology and Neuroscience, University of Glasgow, UK. 

This lead to $39 \times 10 = 390$ interactions each comprising of at least 3 conversational segments that we were able to use to train our models. 
Once the dataset was collected the videos were segmented by hand to isolate the sections that contained only the participants' speech. Most videos contained three speech segments comprised of the participants' answers to each of Pepper's questions. However, some participants followed up on Pepper's responses to their answers resulting in a number of additional speech segments that we were able to add to the corpus. Each of the segments was then labelled by an experimenter in accordance to the self-disclosure score that each participant had assigned to their respective interaction instances. This lead to a total of 1,248 speech and video segments that were used in our deep learning experiments. 

\section{Feature Extraction}\label{FeaturesExtraction}
\subsection{Visual Features}
\begin{figure}
    \centering
    \includegraphics[width=.6\linewidth]{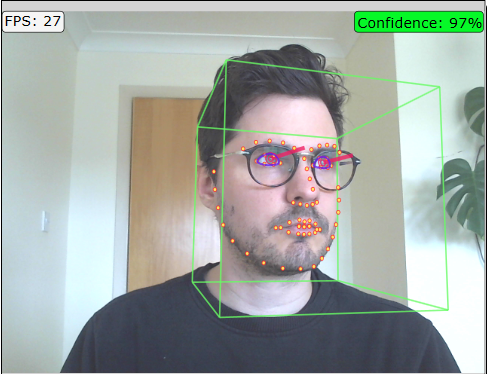}
    \caption{\footnotesize OpenFace 2.2 processing facial action units, and gaze from an input video.}
    \label{fig: openface}
\end{figure}

We extracted a number of visual feature types using a combination of state-of-the art feature extraction models. First, we extracted frame-by-frame gaze and action unit features using the OpenFace 2.2 library \cite{Baltruaitis2018} (see Figure \ref{fig: openface} for visual example). To account for missing frames in each time series that came about as a result of the OpenFace models not registering the presence of a human face, we interpolated the missing frames with the recorded data using spline interpolation. We then filtered and smoothed the resulting multivariate time series with a Savitsky-Golay filter (using a sliding window of 11 frames and a polynomial order of 3). To test the the affects of smoothing and filtering on the results we treated smoothed/filtered and non-smoothed/filtered feature sets as separate in our initial experiments.

\begin{figure}
    \centering
    \includegraphics[width=.6\linewidth]{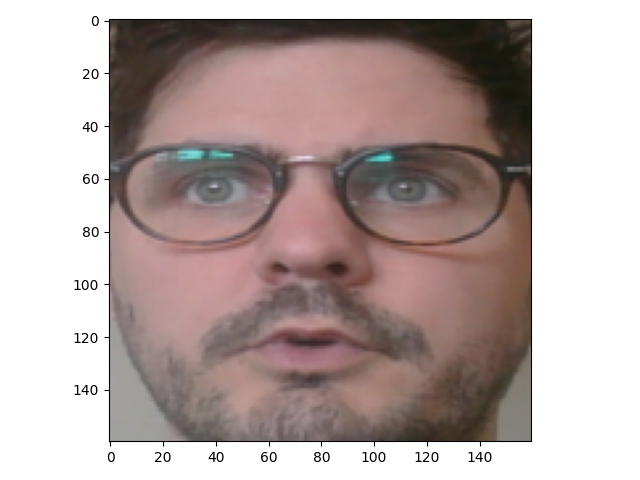}
    \caption{\footnotesize Example output of the MTCNN used for facial feature extraction: a 160x160 pixel normalised face image.}
    \label{fig: mtcnn}
    \end{figure}

Next, we extracted facial features using an InceptionV1 ResNet \cite{Szegedy2015}\cite{He2016-resnet} architecture pretrained on the VGGFace2 dataset \cite{Cao2017VGGF}. VGGFace2 consists of 3.31 million images of celebrity faces organized into 9131 subject categories with large variances in pose, age, illumination, and ethnicity. The InceptionV1 ResNet that we used scored an accuracy of 99.6\% on this dataset. Pre-processing of the video frames in this case consisted of extracting a 160x160 pixel sub-region of each frame that contained pixel and feature-wise normalization of the subject's face (an example of the MTCNN output can be seen in Figure \ref{fig: mtcnn}). This was done using a pretrained multi-task cascaded convolutional neural network \cite{Zhang2016MTCNN} on each video frame. The pretrained ResNet produces 512 facial features for each video frame. Similar to our approach with the the OpenFace features we interpolated and filtered the resulting time series to experiment with the effects that this would have the models' scores.


\subsection{Audio Features}
For audio features we first produced a mel-filter cepstral coefficient (MFCC) matrix for each video's audio modality. This was done using PyTorch's audio feature extraction library using 256 mel-filter banks. This feature set was chosen due MFCCs well established ability to capture significant audio features for human speech recognition tasks \cite{Yang2020recognize}\cite{pawar2021convolution}\cite{kumaran2021fusion}. This was an extension from previous work on using log-mel spectrograms to recognize subjective self-disclosures \cite{Powell2022}, where we found that spectrogram features were more effective at capturing significant self-disclosure related features from subjects' speech. In the case of this study, we found that MFCC features produced better results at initial testing and accordingly we went with MFCC features over the log mel spectrogram alternative. We also experimented with the effects of cepstral mean and variance normalization of MFCC features on our baseline models' performance (detailed in Section \ref{sec:SVMBaseline} as this was a factor that would also have to be taken into account when training our deep learning models).

Second, we extracted audio features directly from each sound file's amplitude array using Facebook AIs wav2vec2.0 architecure \cite{Baevski2020Wav2Vec2}. Wav2vec2.0 uses a stack of convolutional neural network based feature encoders and generates contextualised audio representations using a transformer model \cite{Vaswani2017_transformers}. We used a wav2vec2.0 model pretrained on 960 hours of unlabelled audio data from the LibriSpeech dataset \cite{Panayotov2015libri}. To get the feature sets for each wav file we took the outputs from the models 12 transformer layers which resulted in 12 $t$ x 768 feature matrices where the value t was determined by the number of frames in the audio file.

\section{Deep Learning Experiments}\label{DeepLearningExperiments}
\subsection{Support Vector Machine Baselines}\label{sec:SVMBaseline}
Since we were working with a novel dataset designed specifically for our deep learning experiments we needed some way of establishing a baseline that we were able to compare our results to. Following \cite{Lin2021} we used Gaussian kernel support vector machines (SVM) trained on our extracted audio and visual features separately to establish such a baseline. For each feature type, a vector representing the mean over all frames in each example was computed and the SVMs were tasked with classifying the self-disclosure score for each interaction. Each model was trained using 3 fold cross validation and the average f1 score was used as a means to measure the overall performance of each model. 

\begin{figure}
 \centering
   \includegraphics[width=0.75\columnwidth]{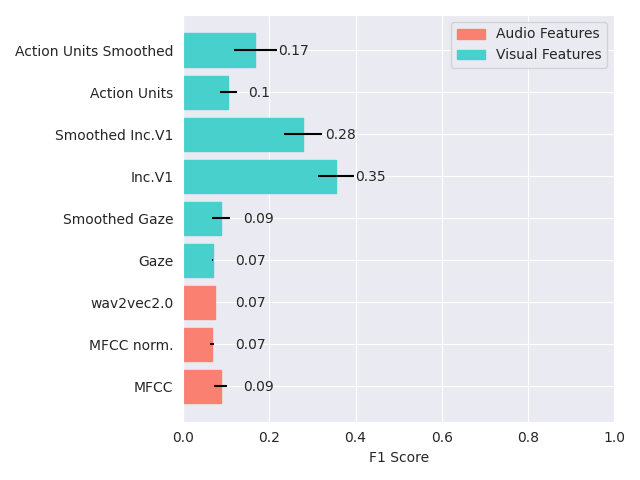}
   \caption{\footnotesize Gaussian SVM baseline F1 scores for individual smoothed/filtered and unsmoothed/unfiltered audio and visual feature sets. Standard deviation is represented by black error bars.}
   \label{SVMbaselinesplot}
\end{figure}

The results of these baseline experiments (illustrated in Figure \ref{SVMbaselinesplot}) indicate that the facial features extracted using InceptionV1 pretrained on VGGFace2 were significantly the most informative for the task while for the audio features, the MFCC representation was the most informative. Overall video features were the most useful feature sets in discrmiitinating the self-disclosure score classes. The results also show that the problem is a difficult one given that the best f1 score measured was only 0.36. One surprising result was that the word2vec2.0 features performed so poorly. We hypothesised that, given the strong relative performance of the InceptionV1 features, that word2vec2.0 would also perform relatively well given that both models are pretrained on large amounts of task relevant data. One possible explanation of why word2vec2.0 features performed so poorly in the baseline test is with respect to how the mean vector for each frame was computed. The word2vec2.0 features for each frame were of far higher dimension than both the MFCC features ($12 \times t^{W}\times178$ vs. $256 \times t^{M}$) and the visual feature set of the highest dimensionality (InceptionV1 features at $t^{I} \times 512$) \footnote{Here $t^{W}$, $t^{M}$, and $t^{I}$ refer to the time dimension of the word2vec2.0 features, the MFCC features, and the InceptionV1 features respectively.}. Thus condensing the word2vec2.0 features across both the time and attention-head dimensions into a single 178 dimensional vector could have meant that too much information was lost leading to the feature dramatically losing its discriminative ability with respect to the task.

\subsection{Multimodal Attention Network}

 \begin{figure*}[h]
 \centering
   \includegraphics[width=.7\textwidth]{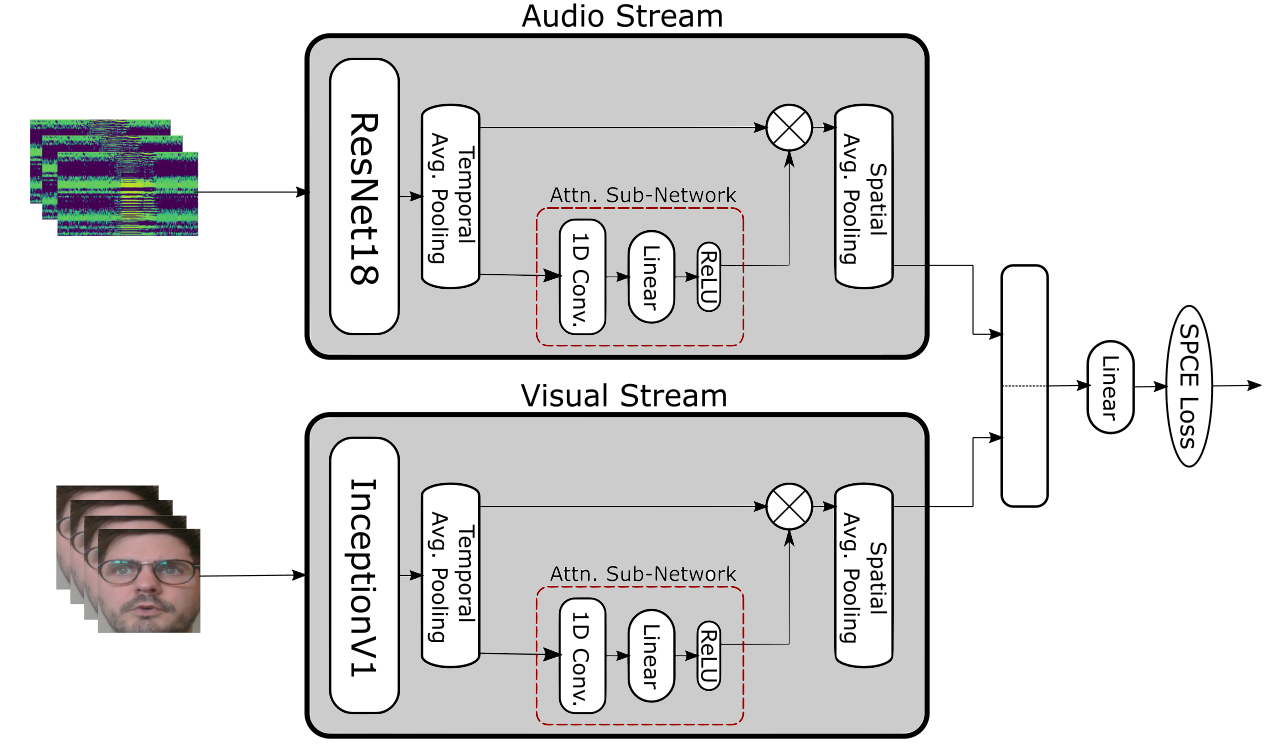}
   \caption{\footnotesize Illustration of our multi-modal attention network. Segments of MFCC matrices (top) and face-cropped video frames (bot) are fed into two similar streams. MFCC segments are fed through an ImageNet pretrained 2DResNet backbone before being average pooled, and cloned. One copy is then sent through the attention subnetwork before being multiplied to the other ResNet output copy. This representation is then average pooled once again producing the final audio embedding. The same process occurs with the frame input except that the backbone is a InceptionV1 ResNet architecture pretrained on VGGFace2. The resulting audio and visual embeddings are then concatenated and fed through a linear classification layer. The network probabilities are then used to compute the scale-preserving cross entropy loss by which the parameters of the network are optimised.}
 \label{fig: attnNet}
\end{figure*}

We designed a multimodal attention network that processes the audio and visual features of each video in separate streams and then combines these representations in a late fusion fashion before being classified by a linear neural network layer. This approach was motivated by previous observations communicated in \cite{Powell2022} that concluded that 'off the shelf' neural network architectures, i.e. ones that were not designed specifically for the task at hand and used no pretraining, produced less than desirable results on the audio-only version of this task. We aimed to improve our previous attempt (i.e., \cite{Powell2022}) by: firstly, taking into account video recordings of the interactions. Secondly, designing a custom neural network architecture that deals with audio and visual features separately before combining them into one latent representation. Thirdly, using pretrained neural network backbones in each feature processing stream and finally, experimenting with feature fusion using principle components analysis to prevent our results from being limited by being only able to use a single feature representation.

The design of this architecture is inspired by other deep learning approaches that utilize attention mechanisms leveraged from deep convolutional neural networks for recognition tasks involving visual and audio data captured from human subjects - specifically in emotion recognition and related tasks \cite{zhao2021combining} \cite{Zhao2020vaa}. Our approach is similar to that in \cite{Zhao2020vaa} in that we use their convolutional architecture for each of the attention mechanisms, although in our case we use only frame-wise attention in both the audio and visual streams. We also use an InceptionV1 ResNet trained on VGGFace2 instead of the 3DResnet used in that study as it was more suited to our problem and our baseline SVM experiment showed convincingly that this feature representation was the most informative for the task. As in \cite{Zhao2020vaa} we compute the frame-wise attention (i.e. along the time dimension in each case) for the audio and visual streams in the following way. We adapt their formulation here for the sake of completeness and clarity with respect to how we have modified their approach. The full architecture is displayed in Figure \ref{fig: attnNet}

\subsubsection{Audio Temporal Attention Subnetwork}
Let $\textbf{x}_{i}^{A}$ be the $i^{th}$ audio feature matrix input. We first center crop $\textbf{x}_{i}^{A}$ to a fixed length $l$ such that $\frac{l}{s} \in \mathbb{N}$ for some positive integer $s$ giving $\textbf{x}_{i}^{A'}$. If the time dimension of $\textbf{x}_{i}^{A}$ is less than $l$ then we pad the input on either side with zeros such that it's length is now equal to $l$. We then split $\textbf{x}_{i}^{A'}$ into $s$ segments and stack them on top of one another such that $\textbf{x}_{i}^{A'} \in \mathbb{R}^{s \times \frac{l}{s} \times n}$. The model then receives a batch of size $b$ of these tensors which is then fed through the model's audio stream. 

The first step of the audio stream is to process each of the $b \times s$ feature segments through a ResNet18 model \cite{He2016-resnet} pretrained on the ImageNet dataset. This may sound surprising given that we are using using an ImageNet trained model on MFCC audio representations (since ImageNet contains no MFCC examples) but research has shown that using such ResNets on MFCC features matrices dependably improves model scores \cite{palanisamy2020rethinking} and indeed we also found this to be the case in our experiments. We then take the output $F^{A}_{j}$ of the fifth convolutional stack of the pretrained ResNet18 model and perform spatial average pooling over the feature maps producing $F^{A'}_{j}$ (where $j$ indexes over the feature matrix segments). This downsamples the output of the ResNet from $F^{A}_{j} \in \mathbb{R}^{s \times h \times w \times c}$ to $F^{A'}_{j} \in \mathbb{R}^{s \times c}$ where $s$ is the number of segments, $h$ and $w$ are the height and width of the feature maps respectively, and $c$ is the number of channels, creating a $1 \times c$ length descriptor for each of the segments. The goal is now to learn an $s \times 1$ length descriptor for the audio feature matrix segments where the $k^{th}$ element of the descriptor weights the $k^{th}$ segment according to its importance in classifying the input sample. This descriptor is learned using a convolutional stack that consists of a 1D convolutional layer, a fully connected linear layer, and a ReLU non-linearity such that:

\begin{align}
     H^{A} &= W^{A}_{1}(W^{A}_{2}(F^{A'}_{j})^{T})^{T}
\end{align}

Where $W^{A}_{1}$ and $W^{A}_{2}$ are $s \times s$ and $1 \times c$ learnable parameter matrices for the linear and convolutional layers respectively. We then compute the activation of the audio attention subnetwork $A^{A}$ i.e. the $s \times 1$ length segment descriptor as:

\begin{align}
    A^{A} &= \text{ReLU}(H^{A})
\end{align}

The output embedding for the audio stream, i.e. the representation of which audio segments are most relevant to the classification of the input example to a particular self-disclosure class, is computed via:

\begin{align}
    E^{A} &= \sum_{j=1}^{S} F^{A'}_{j}  A^{A}
\end{align}

\subsubsection{Visual Temporal Attention Subnetwork}
The approach to achieve the audio embedding $E^{V}$ for the visual features extracted from the videos follows precisely the same steps as the audio temporal attention algorithm. The principal differences in practice are that we use the InceptionV1 ResNet architecture trained on VGGFace2 that we used in our SVM baseline experiments instead of the ResNet18 model.

Given the output embeddings $E^{A}$ and $E^{V}$ for the audio and visual processing streams we then summarize the features using average pooling by computing the mean of each embedding vector along the time domain (i.e. across segments) giving  $E^{A'}$ and $E^{V'}$. These are then concatenated before being fed to a linear layer containing 7 neurons representing each of the self-disclosure score classes. This produces output $\hat{\textbf{y}}$:

\begin{align}
    \hat{\textbf{y}} &= \text{Softmax}(W^{AV} \text{concat}(E^{A}, E^{V}))
\end{align}

where $W^{AV}$ is a learnable parameter matrix related to the linear output layer, $\text{concat()}$ is the concatenation operation, and $\text{Softmax()}$ is the softmax function that return normalized probabilities over the seven self-disclosure score classes.

\subsection{Ablation Experiment Parameters}
In our experiments we tested the influence of two different visual feature sets, two experimental framings, and four different loss functions to determine the best configuration for the problem. 
\subsubsection{Visual Feature Sets}
First, we wanted to test the efficacy of just the facial features output by the InceptionV1 ResNet architecture pretrained on VGGFace2 as our SVM experiments showed that these were likely to be the most informative visual features for the task. Next we wanted to test a combination of all visual features that we extracted as detailed in Section \ref{sec:SVMBaseline}. To reduce the dimensionality of this feature space we concatenated all of the visual features together after the visual input has been passed through the ResNetV1 backbone in the visual stream and performed principal components analysis with parameters set such that 99\% of the variance in the data was explained by the resulting feature matrix. This resulted in a dimensionality change in this feature space from a 555 dimensional feature vector to a 67 dimensional feature vector for each video frame. 

\subsubsection{Classification Vs. Regression}
In \cite{Powell2022} we found that there was a nuance in the approach to classifying self-disclosure scores. As we state in that study, participants rated the degree of self-disclosure in their interactions on a likert scale between 1 and 7. This means that each score falls into a discrete class meaning that one plausible way to frame the problem is as an n-class classification problem. However, loss functions related to n-class classification problems often treat incorrect guesses in the same manner i.e. there is no sense in which one guess can be numerically represented as being closer to a correct guess than any of the other possible guesses. The self-disclosure score data, however, is scaled in the sense that a model guess of 2 for ground truth self-disclosure score of 1 should be treated as a better guess than 6 or 7. In this light an argument could be made that the problem is better represented as a regression problem. In \cite{Powell2022} we found that framing the problem in both ways produced similar results and as such no clear empirically informed decision could be made about what approach worked best. In light of this we decided to test the effects of both approaches on our results. 

\subsubsection{Loss Function}
We wanted to study the effect of loss function on the problem. Standardly, regression based methods minimize a mean-squared error loss in order to optimize the parameters of a given model. Since we had no good reason to suspect that this particular problem required an alternative regression-based loss function we chose only to base our regression results on the mean squared error loss. For the classification version of the problem we chose a categorical cross-entropy loss function for our experiments. For this loss, research has shown that label smoothing, a technique whereby standard 'hard' labels are modified by a smoothing parameter $\alpha$ via $y_{k}^{LS}=y_{k}(1-\alpha)+\frac{\alpha}{K}$ where $k$ indexes over the total number of classes (seven in the case of this study), can drastically improve results \cite{yuan2020revisiting}. As such we chose to include a cross entropy loss with label smoothing as part of ablation study. Last, we wanted to explore the possibility of designing a custom loss function that was able to strike a balance between the classification and regression versions of the task i.e. one that leveraged the fact that the data was categorical while also preserving the notion that certain guesses were better with respect to a ground truth label than others. Taking inspiration from \cite{Zhao2020vaa} we designed a custom cross entropy loss function that penalises guesses with greater severity the further they are from the ground truth label. For example, for an input sample with labelled self-disclosure score of 7 a guess of 1 will result in a higher loss than a guess of 2, a guess of 2 will result in a higher loss than a guess of 3, and so on. To do this we amended the standard cross-entropy loss function which can be expressed as:

\begin{align} \label{CELoss}
    \mathcal{L}_{CE} = -\frac{1}{N}\sum_{i=1}^{N}\sum_{c=1}^{C}\mathds{1}_{[c=y_{i}]}\log p_{i, c}
\end{align}

where N is the number of input samples, C is the number of classes, $\mathds{1}_{[c=y_{i}]}$ is an indicator variable that equals 1 when the predicted class is the same as the ground truth class and $p_{i, c}$ is the probability that the $i^{th}$ sample belongs to the $c^{th}$ class. We added a penalty term to \ref{CELoss} that formalizes the idea that guesses at a greater distance from the ground truth should be penalised more severely. This gives what we term the scale preserving cross-entropy loss:

\begin{align} \label{SPCELoss}
    \mathcal{L}_{SPCE} = -\frac{1}{N}\sum_{i=1}^{N} (1+\lambda(|y-\hat{y}|)^{\mu}) \sum_{c=1}^{C}\mathds{1}_{[c=y_{i}]}\log p_{i, c}
\end{align}

where $\lambda$ and $\mu$ are hyperparamters that change the degree to which a incorrect guess is penalised with respect to how far away it is from the ground truth self-disclosure label.

Taken together, the parameters of the ablation study lead to eight different training configurations for the multimodal attention network, the specifications of which are presented in Figure \ref{ResultsFig}.

\subsection{Model Training}
Regression and classification models were trained over 100 epochs, while the SPCE models where trained on 150 since we found that they took longer to converge. All network version we trained using the Adam optimizer \cite{Kingma2014}, an initial learning rate of 0.01, and mini-batch size of 35. Audio feature inputs were cropped to length $l=128$ and divided in to $s=4$ segments. Visual input features were cropped to $l=210$ frames and divided into $s=7$ segments. We prepared the training data as in \cite{Powell2022} splitting the training and testing datasets into an 80/20 split and used weighted random sampling to account for imbalanced classes. Each model was trained five times and the average F1 score and standard deviation over all five training instances were computed to give a balanced assessment of the model's performance. We chose to validate the models using f1 scores so that our results were directly comparable to those produced by our SVM experiments.

\section{Results}\label{Results}
\begin{figure}
 \centering
   \includegraphics[width=0.75\columnwidth]{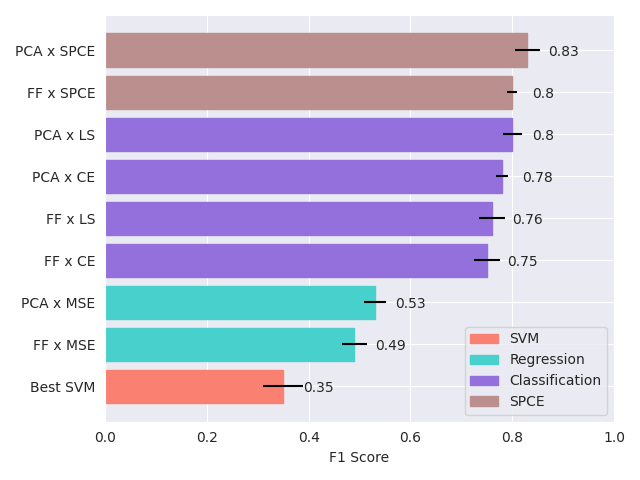}
   \caption{\footnotesize F1 scores for our multimodal attention network trained on a different combination of data input representations (principal components analysis data (PCA), face features only (FF)) and loss functions (categorical cross entropy (CE), cross entropy with label smoothing (SE), mean squared error (MSE), and our scale preserving cross entropy loss (SPCE)). We have also colour coded the different experimental framings we used for the deep learning experiments.}
    \label{ResultsFig}
\end{figure}

The results of our  study are displayed in \ref{ResultsFig}. We found that all versions of the multimodal attention network scored significantly above the best SVM baseline. Interestingly, departing from \cite{Powell2022}, where regression and classification models performed about as well as each other, we found that a classification framing (treating self-disclosure scores as discrete classes) was significantly more effective at modelling the problem than a regression framing (treating the scores as being derived from the continuous number line). In all cases we found that, within each experimental framing, the features derived from principle components analysis outperformed models trained on just InceptionV1 facial features. This is perhaps unsurprising for two reasons. Firstly, because this feature set was comprised of three times the number of features than the pure InceptionV1 feature set before it was condensed to its principal components. Secondly, because significantly reducing the number of features (from 512 in the pure InceptionV1 case to 67 in the principal components case) would mean that our model was less susceptible to the curse of dimensionality i.e. that it would require much less data to effectively model that smaller set of features. Further, we found that label smoothing produced improvements in results when compared to the non-label smoothing variant of the cross entropy loss. Finally, we found that our scale preserving cross-entropy loss outperformed all but one version of the model (principal component features with label smoothing cross-entropy loss) to which it equalled in performance.

\section{Discussion and Conclusion}\label{DiscussionConclusion}
Overall we report significant increases on performance in this task from previous attempts (e.g., \cite{Powell2022}). We hypothesise that this is due to a number of significant developments from that work. Firstly, we collected a much larger dataset meaning that the models had more examples to learn from. Second, in this case all interactions were recorded between the same interaction dyad (i.e. between a human and a Pepper robot). In \cite{Powell2022}, we collected interaction data between three different dyads: human--human, human--embodied robot, and human--voice agent. One reason that results may have been worse in that case is due to the possibility that vocal features particular to each self-disclosure class may have been modulated by the kind of agent the participant was interacting with. Since the interaction partner remained constant in our study, variability was reduced, which could have simplified the learning task. Moreover, we used significantly more advanced models, taking advantage of the representational power of large deep neural networks trained on extensive datasets. Further, our use of frame-wise attention mechanisms make use of deep learning techniques that have shown to be state-of-the art on video and language modelling tasks perhaps providing a straightforward upgrade of the 'off-the-shelf' models that were used in the previous study. Lastly, and perhaps most obviously, in this study we modelled two different sensory modalities (audio and visual) as opposed to the single sensory modality that was considered in \cite{Powell2022}. It may well be the case that the auditory domain holds less discriminative information than the visual domain for self-disclosure modelling and thus, the previous study was automatically at a disadvantage in only considering the former.

One surprising observation from our baseline experiments was that the visual features were most effective at allowing the SVM models to predict a particular subjective self-disclosure score. While much of the literature on self-disclosure is varied with respects to its definitions one thing that is generally agreed upon is that self-disclosure is primarily a verbally communicated social phenomenon \cite{cozby1973self,RefWorks:455}. In light of this it might be expected that the audio modality would produce the best results. It may well be the case that the way in which the audio features were averaged caused some of the information to be lost. Unfortunately, a thorough investigation of why the visual features were the most informative is outside of the scope of this paper.

While this study shows significant improvement on previous work done on modelling self-disclosure with neural networks, it remains to be seen whether the advances that we detail here are significant enough for these models to be implemented in social robots. There is a significant risk associated with an incorrect self-disclosure scoring in a real world setting \cite{Powell2022}. Assuming that a person is sharing very little self-disclosure when in fact they believe themselves to be sharing a significant amount could lead to that person feeling as if they are being ignored or that the sensitive information that they are sharing is not worthy of the listener's consideration \cite{Collins1994}. Conversely, assigning a very high self-disclosure score in a situation where an interaction partner does not believe themselves to be sharing a significant amount of personal information could cause undue levels of attention to be paid to a situation which is not important \cite{Reese2023}. The issue described in both of these cases would be significantly confounded within the context of emotional well being interventions, where the risks associated with not picking up on a user's self-disclosure related signals could be very damaging. As such, considerably more work needs to be done before models like ours are considered for real world application. There are at least two ways that steps could be taken in this direction. Firstly, significantly more data should be collected to improve the performance of the models. Secondly, a study should be carried out to asses the differences between model performance and the performance on the same task by a trained professional. It is often the case that the quality of a machine learning model and it's viability as a real world application is measure with respect to its ability to achieve 'human-like' performance. It makes sense that a model that is effective at recognizing the degree to which a person is disclosing personal information should be able to do so at least as well as a trained professional (particularly if that model is to be implemented within the context of health care interventions).

Further, there are ways in which improvements on our approach might be made in the short term. Firstly, since we found that performance on the task was improved when visual features were combined using principal components analysis, it's likely to also be the case that performance improvements could be achieved by combining audio features. In particular, we did not experiment with ways to combine outputs from the transformer layers of wave2vec2.0 with the MFCC features. Additionally, more empirical work could be done to ascertain the best way to combine feature sets in both the audio and visual cases. For one such example, \cite{Lin2021} used a denoising autoencoder to learn a compressed latent representation of the concatenated input features. A future study should empirically test the hypothesis that such a latent representation exists in a more effective input feature space than the one produced by principal components analysis. Further studies could also look into experimenting with other kinds of attention. In \cite{Zhao2020vaa} the authors use channel-wise attention and spatial attention in the visual stream on top of the frame-wise attention that both of our methods share. One development along these lines could be to implement an attention mechanism that produces a descriptor over the features i.e. the columns of the input matrices. In this way the model would hold a representation of not only which frames of the input are important to its classification but also which features are important. Lastly, the model could be altered to leverage 3D ResNets to produce higher dimensional features over the input video frames. This approach however would require a 3DResNet trained on a very large video dataset focused on the modelling of human faces and, to our knowledge, no such pretrained model is publicly available. Taking from the modelling literature on self-disclosure \cite{soleymani2019multimodal}, show that very good results on the task of (non-subjective) self-disclosure modelling between two human interactants can be achieved multi-modally with the addition of lexical features. In that study, the authors use a pretrained BERT language model \cite{devlin-etal-2019-bert} to extract features related to the words used in each utterance. A significant part of self-disclosure (at least in the human--human case) is thought to be communicated verbally \cite{cozby1973self}\cite{RefWorks:455}. This a future study could look at including this modality in the HRI version of the task.

We believe that this study makes significant strides into the new field of subjective self-disclosure modelling by showing considerable improvements over results of any previous studies on the topic. 

\bibliographystyle{ieeetr}
\balance{ \bibliography{refyr}}

\end{document}